\documentclass[journal]{IEEEtran}
\usepackage{graphicx}
\usepackage{colortbl}
\usepackage[table*]{xcolor}
\usepackage{algorithm}
\usepackage{algorithmic}
\usepackage[hyphens]{url}  
\usepackage{multirow}
\usepackage{multicol}
\usepackage{amssymb}
\usepackage{amsmath}
\usepackage{cite}
\usepackage{caption}
\usepackage{amsmath}
\usepackage{amssymb}
\usepackage{bm}
\usepackage[colorlinks,linkcolor=red]{hyperref}
\usepackage{multirow}
\usepackage{graphicx}  
\usepackage{caption} 
\begin{document}
\title{Quantized Neural Networks via $\{-1, +1\}$ Encoding Decomposition and Acceleration}

\author{
        Qigong~Sun,~\IEEEmembership{Student Member,~IEEE,}
        Xiufang~Li,
        Fanhua~Shang,~\IEEEmembership{Senior Member,~IEEE,}
        Hongying~Liu,\\
        Kang~Yang,
       Licheng~Jiao,~\IEEEmembership{Fellow,~IEEE,}
       and~Zhouchen~Lin,~\IEEEmembership{Fellow,~IEEE}
\thanks{Preliminary results of this paper appeared in our previous conference paper \cite{sun2019qnn}.}
\IEEEcompsocitemizethanks{	
\IEEEcompsocthanksitem Q.\ Sun, X.\ Li, F.\ Shang, Y.\ Liu, K.\ Yang, and L.\ Jiao are the Key Laboratory of Intelligent Perception and Image Understanding of the Ministry of Education,
International Research Center for Intelligent Perception and Computation, Xidian University, Xi'an Shaanxi Province 710071, China. E-mails: xd\_qigongsun@163.com; xfl\_xidian@163.com; \{fhshang, hyliu\}@xidian.edu.cn; yangk12321@gmail.com; lchjiao@mail.xidian.edu.cn.
\IEEEcompsocthanksitem Z.\ Lin is with Key Laboratory of Machine Perception (MOE), School of EECS, Peking University, Beijing 100871, P.R.\ China. E-mail: zlin@pku.edu.cn.
}
}

\maketitle

\IEEEtitleabstractindextext{%
\begin{abstract}
The training of deep neural networks (DNNs) always requires intensive resources for both computation and data storage. Thus, DNNs cannot be efficiently applied to mobile phones and embedded devices, which severely limits their applicability in industrial applications. To address this issue, we propose a novel encoding scheme using $\{-1, +1\}$ to decompose quantized neural networks (QNNs) into multi-branch binary networks, which can be efficiently implemented by bitwise operations (i.e., \emph{xnor} and \emph{bitcount}) to achieve model compression, computational acceleration, and resource saving. By using our method, users can achieve different encoding precisions arbitrarily according to their requirements and hardware resources. The proposed mechanism is highly suitable for the use of FPGA and ASIC in terms of data storage and computation, which provides a feasible idea for smart chips. We validate the effectiveness of our method on large-scale image classification (e.g., ImageNet), object detection, and semantic segmentation tasks. In particular, our method with low-bit encoding can still achieve almost the same performance as its high-bit counterparts.
\end{abstract}

}

\IEEEdisplaynontitleabstractindextext
\IEEEpeerreviewmaketitle

\section{Introduction}\label{sec:introduction}
\IEEEPARstart{D}{eep} Neural Networks (DNNs) have been successfully applied in many fields, such as image classification, object detection and natural language processing. Because of a great number of parameters in multilayer networks and complex model architectures, huge storage space and considerable power consumption are needed. For instance, the VGG-16 model has a total of $138$ million parameters, and requires more than $500$MB storage space and $30$ billion float point operations in the inference process. With the rapid development of chip (e.g., GPU and TPU) technologies, computing power has been dramatically improved. In the rapid evolution of deep learning, most researchers use multiple GPUs or computer clusters to train deeper and even more complex neural networks. Nevertheless, the energy consumption and limitation of computing resources are still significant factors in industrial applications, which are generally ignored in scientific research. In other words, breathtaking results of many DNN algorithms under the condition of applying GPUs are lagging behind the demand of the industry. DNNs can hardly be applied in mobile phones and embedded devices directly due to their limited memory and calculation resources. Therefore, the model compression and acceleration in DNNs are especially important in the future industrial and commercial applications.

In recent years, many solutions have been proposed to improve the energy efficiency of hardware and achieve model compression or computational acceleration, such as network sparsification and pruning \cite{Hassibi1993Second,Wen2016Learning,Tran2015Learning}, low-rank approximation \cite{denton2014exploiting,jaderberg2014speeding,tai2015convolutional},
designing architectures \cite{Howard2017MobileNets,Sandler2018MobileNetV2} and model quantization \cite{hubara2017quantized,Rastegari2016XNOR,lin2017towards}.
Network sparsification and pruning can dramatically decrease the number of connections in the network, and thus considerably reduce the computational load in the inference process without a pretty high loss in accuracy. Luo \emph{et al.} \cite{LuoThiNet} presented a unified framework, and exploited a filter level pruning method to discard some less important filters. \cite{tai2015convolutional} used low-rank tensor decomposition to remove the redundancy in convolution kernels, which can serve as a generic tool for speed-up.
Since there is some redundant information in the networks, most straightforward approaches of cutting down those information are to optimize the structure and yield smaller networks \cite{Ioffe2015Batch,Iandola2016SqueezeNet}. For example, \cite{Howard2017MobileNets} proposed to use bitwise separable convolutions to build light networks for mobile applications. Most of those networks still utilize floating-point representations (i.e., full-precision values). However, \cite{gupta2015deep} discussed that the full-precision representations of the weights and activations in DNNs are not necessary during the training, and a nearly identical or slightly better accuracy may be obtained under lower-precision representation and calculation. Thus, many scholars are working on the topic of model quantization to achieve model compression and resource saving. Model quantization can be applied to many network architectures and applications. In particular, it can also combine with other strategies (e.g., pruning) to achieve more efficient performance.

Model quantization methods mainly include low-bit quantization approaches (e.g., BNN \cite{Courbariaux2016Binarized}, XNOR-Net \cite{Rastegari2016XNOR}, TWN \cite{Li2016Ternary}, and TBN \cite{wan2018tbn}) and $M$-bit (e.g., $M\!=\!4$ or $M\!=\!8$) quantization approaches (e.g., DoReFa-Net \cite{Zhou2016DoReFa}, INQ \cite{Zhou2016Incremental}, ABC-Net \cite{lin2017towards}, and LQ-Nets \cite{Zhang2018LQ}). Those low-bit quantization methods can achieve extreme model compression, computational acceleration and resource saving. For example, XNOR-Net \cite{Courbariaux2016Binarized} uses bitwise operations (i.e., \emph{xnor} and \emph{bitcount}) to replace full-precision matrix multiplication, achieving 58$\times$ speedups and 32$\times$ memory saving in CPU \cite{Rastegari2016XNOR}. As discussed in \cite{Liang2017FP}, FP-BNN attains a higher acceleration ratio on FPGA, which can speed up to about 705$\times$ in the peak condition compared with CPU, and is 70$\times$ faster than GPU. However, most models were proposed for a fixed precision, and cannot be extended to other precision models. The representation capability of low-bit parameters is insufficient for many practical applications, especially for large-scale image classification (e.g., ImageNet) and regression tasks. Therefore, the low-bit quantization methods suffer from significant performance degradation. In contrast, $M$-bit quantization networks have better representation capability than low-bit quantization ones, and thus can be applied to more complex real-world applications. Some scholars use $M$-bit quantization to improve both accuracy and compression ratio of their networks, but seldom consider their computational acceleration \cite{gupta2015deep, vanhoucke2011improving, Zhou2016Incremental}. Both ABC-Net \cite{lin2017towards} and LQ-Nets \cite{Zhang2018LQ} used the linear combination of multiple binary parameters constrained to $\{-1, +1\}$ to approximate full-precision weights and activations. Therefore, the complex full-precision matrix multiplication can be decomposed into some simpler operations. \cite{Guo2017Network} and \cite{Xu2018Alternating} used the same technique to accelerate the training of convolutional neural networks (CNNs) and recurrent neural networks (RNNs). However, those methods not only increase the number of parameters many times, but also introduce a scale factor to transform the original problem into an NP-hard problem, which makes the solution difficult and highly complex.
The slow convergence rate in training process and the lack of efficient acceleration strategies in inference process are two typical issues in $M$-bit quantization networks.

In order to extend the efficient bitwise operations (i.e., \emph{xnor} and \emph{bitcount}) of binary quantization to $M$-bit quantization without excessively increasing time complexity and space complexity, we propose a novel encoding scheme using $\{-1, +1\}$ to decompose trained quantized neural networks (QNNs) into multi-branch binary networks.
Our method bridges the gap between binary and full-precision quantizations, and can be applied to many cases (from $1$-bit to $8$-bit). Thus, our encoding mechanism can improve the utilization of hardware resources, and achieve parameter compression and computation acceleration. In our experiments, we not only validate the performance of our multi-precision quantized networks for image classification on CIFAR-10 and large-scale datasets (e.g., ImageNet), but also implement object detection and semantic segmentation tasks.
The main advantages of our method are summarized as follows:
\begin{itemize}
  \item \textbf{Training easier}. For efficiently training our networks, we use the parameters of high-bit trained models sequentially to initialize a low-bit model,
    which converges faster than those straightforward approaches (e.g., DoReFa-Net, ABC-Net, and INQ).
    Hence, our networks can be trained within a short time, and only dozens of times fine-tuning is needed to achieve the accuracies in our experiments.
      Thus, our multi-branch binary networks can be easily applied to various practice engineering applications.
  \item \textbf{More options}. Different from BWN, BNN, XNOR-Net and TWN, our method can provide $64$ available encoding options for different encoding precision networks. Naturally, each option with a special encoding precision has a different calculation speed, resource requirement and experimental precision. Therefore, users can choose the appropriate option for their requirements.
  \item \textbf{Suitable for hardware implementations}. After the process of decomposition, instead of storing all encoding bits in data types, e.g., char, int, float or double, the parameters can be individually stored by bit vectors. Thus, the smallest unit of data in electronic equipments can be reduced to $1$-bit from $8$-bit, $16$-bit, $32$-bit or $64$-bit, which raises the utilization rate of resources and compression ratios of the model. Then the data can be encoded, calculated and stored in various encoding precisions.
  \item \textbf{Suitable for diverse networks and applications}. In our experiments, we evaluate our method on many widely used deep networks (e.g., AlexNet, ResNet-18, ResNet-50, InceptionV3, MobileNetV2 and DeepLabV3) and real-world applications (e.g., image classification, object detection and semantic segmentation). Our encoding scheme could also be applied to many other networks and applications.
\end{itemize}

\section{Related Work}
DNN quantization has attracted a lot of attention to promote the industrial applications of deep learning, especially for mobile phones, video surveillance, unmanned aerial vehicle and unmanned driving. Many researchers use quantitative methods to get lighter and more efficient networks. All the research mainly focuses on the following three issues.

How to quantize the values of weights or activation in DNNs to achieve model compression?
Quantization methods play a significant role in QNNs, and can be used to convert their floating-point weights and activations into their fixed-point (e.g., from $1$-bit to $16$-bit) counterparts for  achieving model compression. 
\cite{gupta2015deep} used the notation of integer and fractional bits to denote a $16$-bit fixed-point representation, and proposed a stochastic rounding method to quantify weights and activations. \cite{vanhoucke2011improving} used $8$-bit quantization to convert weights into signed char and activation values into unsigned char, and all the values are integer. For multi-state quantification (from $2$-bit to $8$-bit), linear quantization is used in \cite{hubara2017quantized,wang2018two,zhuang2018towards}. Besides, \cite{miyashita2016convolutional} proposed logarithmic quantization to represent data and used bitshift operations in log-domain to compute dot products. For ternary weight networks \cite{Li2016Ternary}, the weights are quantized to $\{-\Delta^*,0,\Delta^*\}$, where $\Delta^*\!\!\approx\!\!\frac{0.7}{n}\!\sum^n_{i=1}\!|W_i|$, $W$ is the model weights and $n$ is its number. In \cite{Zhu2016Trained}, the positive and negative states are trained together with other parameters. When the states are constrained to $1$-bit, \cite{Courbariaux2016Binarized} applied the sign function to quantize both weights and activations to $-1$ and $1$. \cite{Rastegari2016XNOR} also used $\{-\alpha^*, \alpha^*\}$ to represent the binary states, where $\alpha^*\!=\!\frac{1}{n}\sum^n_{i=1}\!|W_i|$.

How to achieve computational acceleration in the inference process? After quantized training, weights and/or activation values are represented in a low-bit form (e.g., $1$-bit and $2$-bit), which has the potential of acceleration computation and memory saving.
The most direct quantization is to convert floating-point parameters into corresponding fixed-point (e.g., $8$-bit) ones, and we can achieve hardware acceleration for fixed-point based computation \cite{gupta2015deep,vanhoucke2011improving,jacob2018quantization,geng2019dataflow}. When the weight is extremely quantized to 1-bit (i.e., binary or $\{-1, +1\}$) as in \cite{Courbariaux2015BinaryConnect} or 2-bit (i.e., ternary or $\{-1, 0, +1\}$) as in \cite{Li2016Ternary,wang2017fixed}, the matrix multiplication can be transformed into full-precision matrix addition and subtraction to accelerate computation. Especially when the weight and activation values are binarized, matrix multiplication operations can be transformed into highly efficient \emph{xnor} and \emph{bitcount} operations \cite{Courbariaux2016Binarized,Rastegari2016XNOR}.
Bitwise operations are based on $1$-bit units, and thus they cannot be directly applied to multi-bit computation. \cite{Guo2017Network,lin2017towards} used a series of linear combinations of $\{-1, +1\}$ to approximate the parameters of full-precision convolution models, and then converted floating point operations into multiple binary weight operations to achieve model compression and computation acceleration. Miyashita \emph{et al.} \cite{miyashita2016convolutional} applied logarithmic data representation on activations and weights that transform the dot product from linear domain into log-domain. Therefore, the dot product can be computed by the bitshift operation to achieve computational acceleration.

How to train QNNs in the quantized domain?
It is clear that some discrete functions, which may be non-differentiable or have zero derivatives everywhere, are required to quantize weights or activation values in QNNs. The traditional stochastic gradient descent (SGD) methods are unsuitable to train the deep networks. Therefore, many researchers are devoting themselves to addressing this issue. All the optimization methods can be divided into two categories: quantizing pre-trained models with or without retraining such as \cite{lin2016fixed,zhou2017incremental,Li2016Ternary,lin2017towards} and directly training quantized networks such as \cite{Courbariaux2015BinaryConnect,Courbariaux2016Binarized,Rastegari2016XNOR,wang2018two}.
\cite{hubara2017quantized}, \cite{Courbariaux2016Binarized,feng2020convolutional}, and \cite{Xu2018Alternating} use the strategy of straight-through estimator (STE) \cite{Bengio2013Estimating} in training process to optimize deep networks. STE uses nonzero gradient to approximate the function gradient, which is not-differentiable or whose derivative is zero, and then applies SGD to update the parameters. Zhuang \emph{et al.} \cite{zhuang2018towards} proposed a two-stage optimization procedure, which quantizes the weights and activations in a two-step manner.
\cite{mishra2017apprentice} and \cite{polino2018model} use high-precision teacher networks to guide low-precision student networks to improve performance, which is called knowledge distillation. \cite{lin2017towards,Zhang2018LQ,hu2018hashing}, and \cite{hu2018training} apply alternating optimization methods to train network weights and their corresponding scaling factors.

Recently, the research of model quantization has appeared mixed-precision quantization \cite{cai2020rethinking,nandakumar2018mixed,uhlich2019mixed,sun2021effective},
robustness quantization \cite{lin2019defensive,shkolnik2020robust,sun2021one}, multiscale quantization \cite{sun2021mwq} and post quantization training \cite{nagel2020up,li2021brecq}.

\section{Multi-Precision Quantized Neural Networks}
\label{gen_inst}
In this section, we propose a novel encoding scheme using $\{-1, +1\}$ to decompose QNNs into multi-branch binary networks. In each branch binary network, we use $-1$ and $+1$ as the basic elements to efficiently achieve model compression and forward inference acceleration for QNNs. Different from fixed-precision neural networks (e.g., binary and ternary), our method can yield multi-precision networks and make full use of the advantage of bitwise operations to accelerate training of our network.

\subsection{Computational Decomposition}
As the basic computation in most layers of neural networks, matrix multiplication costs lots of resources and is also the most time-consuming operation. Modern computers store and process data in binary format, thus non-negative integers can be directly encoded by $\{0, 1\}$. Therefore, we propose a novel decomposition scheme to accelerate matrix multiplication as follows: Let $x\!=\![\mathrm{x}^1,\mathrm{x}^2,...,\mathrm{x}^N]^T$ and $w\!=\![\mathrm{w}^1,\mathrm{w}^2,...,\mathrm{w}^N]^T$ be two vectors of non-negative integers, where $\mathrm{x}^i,\mathrm{w}^i \!\in\! \{0,1,2,...\}$ for $i\!=\!1,2,...,N$. The dot product of those two vectors is written as follows:
\begin{equation}
\begin{split}
x^T\cdot w =&\: [\mathrm{x}^1,\mathrm{x}^2,...,\mathrm{x}^N][\mathrm{w}^1,\mathrm{w}^2,...,\mathrm{w}^N]^T \\
=&\,\sum_{n=1}^{N}\mathrm{x}^n\cdot \mathrm{w}^n.
\end{split}
\end{equation}

All of the above operations consist of $N$ multiplications and $(N\!-\!1)$ additions. Based on the above $\{0, 1\}$ encoding scheme, the vector $x$ can be encoded to a binary form using $M$ bits, i.e.,
\begin{eqnarray}\label{equ01}
x\!=\![\overbrace{\mathrm{c}^1_M\:\! \mathrm{c}^1_{M-1}...\:\!\mathrm{c}^1_1}^{\mathrm{x}^1},\overbrace{\mathrm{c}^2_M\:\! \mathrm{c}^2_{M-1}...\:\!\mathrm{c}^2_1}^{\mathrm{x}^2},...,\overbrace{\mathrm{c}^N_M \:\!\mathrm{c}^N_{M-1}...\:\!\mathrm{c}^N_1}^{\mathrm{x}^N}]^{T}.
\end{eqnarray}
Then the right-hand side of the formulation (\ref{equ01}) can be converted into the following form:
\begin{eqnarray}
\begin{bmatrix}
 \mathrm{c}^1_M&  \mathrm{c}^2_M&  \cdots & \mathrm{c}^N_M\\
 \mathrm{c}^1_{M-1}&  \mathrm{c}^2_{M-1}& \cdots & \mathrm{c}^N_{M-1}\\
 \vdots &  \vdots& \cdots & \vdots\\
 \mathrm{c}^1_1&  \mathrm{c}^2_1& \cdots & \mathrm{c}^N_1
\end{bmatrix}
=\begin{bmatrix}
 c_M\\
c_{M-1}\\
 \vdots\\
 c_1
\end{bmatrix},
\end{eqnarray}
where
\vspace{-2mm}
\begin{eqnarray}
\mathrm{x}^i\!&=&\!\sum_{m=1}^{M}2^{m-1}\cdot  \mathrm{c}^i_m, \quad \mathrm{c}^i_m\in \{0, 1\},\\
c_j\!&=&\![\mathrm{c}^1_j,\mathrm{c}^2_j,...,\mathrm{c}^N_j],\;j=1,2,\ldots,M.
\end{eqnarray}
In such an encoding scheme, the number of represented states is not greater than $2^M$. We also encode another vector $w$ with $K$-bit representation in the same way.
Therefore, the dot product of the two vectors can be computed as follows:
\begin{equation}
\begin{split}
x^T\cdot w =&\sum_{n=1}^{N}\mathrm{x}^n\cdot \mathrm{w}^n    \quad\quad\quad\quad\quad\quad\quad\quad\quad\quad \quad\quad\quad\\
=&\sum_{n=1}^{N}\left(\sum_{m=1}^{M}2^{m-1}\cdot \mathrm{c}^n_m\right)\cdot \left(\sum_{k=1}^{K} 2^{k-1}\cdot \mathrm{d}^n_k\right)\\
=&\sum_{m=1}^{M}\sum_{k=1}^{K}2^{m-1}\cdot 2^{k-1}\cdot c_m\cdot d_k^T.
\end{split}
\end{equation}
where ${d}$ is the encoding element of $\mathrm{w}$.
From the above formula, the dot product is decomposed into $M\times K$ sub-operations, in which every element is $0$ or $1$.
Because of the restriction of encoding and without using the sign bit, the above representation can only be used to encode non-negative integers.
However, it is impossible to limit the weights and the values of the activation functions to non-negative integers.

\begin{figure*}
\setlength{\abovecaptionskip}{0.1cm}
\setlength{\belowcaptionskip}{-0.2cm}
\begin{center}
\includegraphics[width=6.0in]{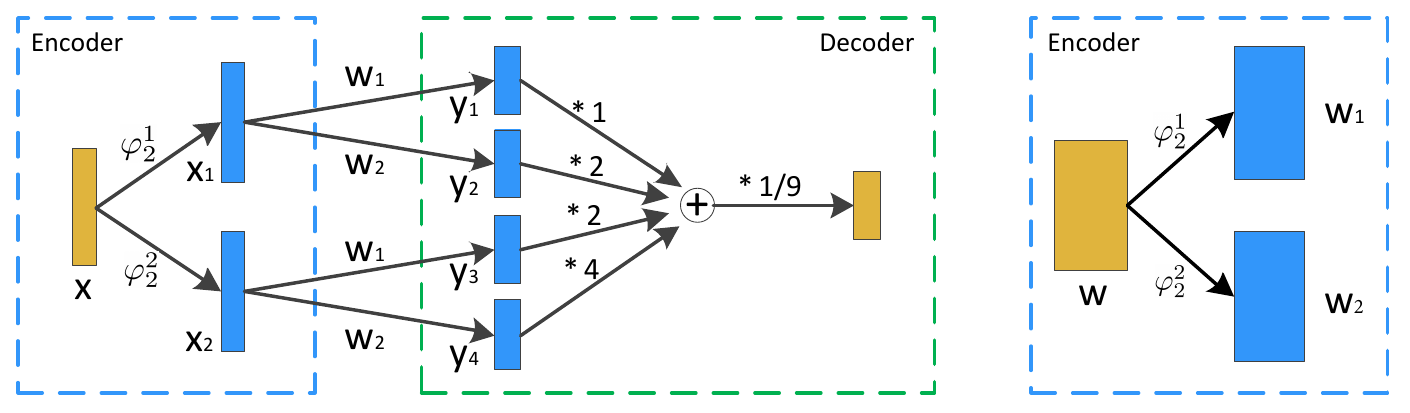}
\end{center}
\vspace{-2mm}

\caption{Our architecture of the 2-bit encoding in a fully connected layer. Here, the 2BitEncoder (i.e., $\varphi _{2}^{1}(x)$ and $\varphi _{2}^{2}(x)$) is used to encode input data and weights in our Encoder stage, and the weighted sum of those four results by fixing scale factors is the final output in our Decoder stage.}
\end{figure*}

In order to extend encoding space to negative integer and reduce the computational complexity, we propose a new encoding scheme, which uses $\{-1, +1\}$ as the basic elements of our encoder rather than $\{0, 1\}$. Except for the difference of basic elements, the encoding scheme is similar to the rules in Eq.\ (4), and is formulated as follows:
\begin{eqnarray}
x^i=\sum_{m=1}^{M}2^{m-1}\cdot \mathrm{c}^i_m, \quad \mathrm{c}^i_m\in \{-1, +1\}
\end{eqnarray}
where $M$ denotes the number of encode bits, and can represent $2^M$ states. And the encoding space is extended to $\{..., -3, -1, 1, 3, ...\}$.
We can use multiple bitwise operations (i.e., \emph{xnor} and \emph{bitcount}) to effectively achieve the above vector multiplications. Therefore, the dot product of those two vectors can be computed as follows:
\begin{eqnarray}
x^T\cdot w =\sum_{m=1}^{M}\sum_{k=1}^{K}2^{m-1}\cdot 2^{k-1}\cdot \emph{XnorPopcount}(c_m, d_k)
\end{eqnarray}
where $\emph{XnorPopcount}(\cdot)$ denotes the bitwise operation, and its details will be defined below. Therefore, this operation mechanism is suitable for all vector/matrix multiplications.

\subsection{Model Decomposition}
In various neural networks, matrix multiplication is one basic operation in some commonly used layers (e.g., fully connected layers and convolution layers). Based on the above computational decomposition mechanism of matrix multiplication, we can decompose the quantized networks into a more efficient architecture.

Let $M$ be the number of encoding bits, and thus the encode space can be defined as: $\{-(2^M\!-\!1), -(2^{M}\!-\!3), ..., (2^{M}\!-\!3), (2^M\!-\!1) \}$, in which each state is an integer value. The boundaries of the values are determined by the number of encoding bits, $M$. However, the parameters in DNNs are usually distributed within the interval $[-r, +r]$. In order to apply our decomposition mechanism to network model decomposition, we use the following formulation for the computation of common layers in neural networks.
\begin{eqnarray}
Op(x, w) = \frac{r}{(2^M\!-\!1)(2^K\!-\!1)}Op(EnC(x), EnC(w))
\end{eqnarray}
where $Op(\cdot)$ is the basic operations for multiple layers (e.g., fully connected operations and convolution operations), $M$ is the number of bits to encode input $x$, $K$ is the number of bits to encode weights $w$, and $EnC(\cdot)$ is the encoding process. Here, ${r}/{[(2^M\!\!-\!1)(2^K\!\!-\!1)]}$ can be viewed as a scaling factor, and $r$ is usually set to $1$. If there is a batch normalization layer after the above layer, the scaling factor is not used in the computation of our decomposition scheme.

For instance, we use a $2$-bit encoding scheme (i.e., $M\!\!=\!\!K\!\!=\!\!2$) for a fully connected layer to introduce the mechanism of our decomposition model, as shown in Fig.\ 1. We define an ``Encoder" that can be used in the 2BitEncoder function (i.e., $\varphi _{2}^{1}(\cdot)$ and $\varphi _{2}^{2}(\cdot)$ defined below) for encoding input data. Here, $x$ is encoded by $x_1\!\in\! \{-1, +1\}^N$ and $x_2\!\in\! \{-1, +1\}^N$, where $x_2$ is high bit data and $x_1$ is low bit data. Therefore, we have the following formula: $x\!=\!x_1 \!+\! 2x_2$. In the same way, the weight $w$ can be converted into $w_1\!\in\! \{-1, +1\}^{M\times N}$ and $w_2\!\in\! \{-1, +1\}^{M\times N}$.
After cross multiplications, we get the four intermediate variables \{$y_1, y_2, y_3, y_4$\}. In other words, each multiplication can be considered as a binarized fully connected layer, whose elements are $-1$ or $+1$. This decomposition can produce multi-branch layers, and thus we call our networks as multi-branch binary networks (MBBNs). For instance, we decompose the $2$-bit fully connection operation into four branches binary operations, which can be accelerated by bitwise operations, and then the weighted sum of those four results by the fixed scaling factor (e.g., $1/9$ in Fig.\ 1) is the final output. Besides fully connected layers, our decomposition scheme is suitable for the convolution and deconvolution layers in various deep neural networks.


\subsection{M-bit Encoding Functions}
As an important part of various DNNs, activation functions can enhance the nonlinear characterization of networks. In our model decomposition method, the encoding function also plays a critical role and can encode input data to multi-bits ($-1$ or $+1$). Those numbers represent the encoding of input data. For some other QNNs, several quantization functions have been proposed. However, it is not clear what the nonlinear mapping is between quantized numbers and encode bits. In this subsection, a list of $M$-bit encoding functions are proposed to produce the element of each bit that follows the rules for encoding data.

\begin{table}[htbp]
\centering
\caption{Activation functions that are used to limit input data to a fixed numerical range.}
\small
\begin{tabular}{c|c}
\hline
\hline
\!\!\!\! $Tanh(x)=\frac{e^x-e^{-x}}{e^x+e^{-x}}$ & $HTanh(x)= \left\{\begin{matrix}
1, & x>1,\\
x, & -1\leqslant x\leqslant 1,\!\!\\
-1, & -1\leqslant x.
\end{matrix}\right.
$ \\
\hline
\!\!\!\! $Sigmoid(x)=\frac{1}{1+e^{-x}}$ & $HReLU(x)= \left\{\begin{matrix}
1, & x>1,\\
x, & 0\leqslant x\leqslant 1,\!\!\\
0, & x\leqslant 0.
\end{matrix}\right.
$ \\
\hline
\hline
\end{tabular}
\end{table}

Before encoding, the data should be limited to a fixed numerical range. Table 1 lists the four activation functions. $HTanh(\cdot)$ is used to project input data to $[-1, +1]$, and it consists of the sign function to achieve binary encoding of weights and activations \cite{Courbariaux2016Binarized,Liang2017FP}. Since the convergence of \emph{SGD} for training the networks with $ReLU(\cdot)$ is faster than other activation functions, we propose a new activation function $HReLU(\cdot)$, which retains the linear characteristics in the specific range and limits the range of input data to $[0, 1]$.
Different from general activation functions mentioned above, the output of our $M$-bit encoding function defined below should be $M$ numbers, which are $-1$ or $+1$. Those numbers represent the encoding of input data. Therefore, the dot product can be computed by Eq.\ (9). In addition, in the above described experimental condition, when we use $2$-bit to encode the data $x$ and constrain to $[-1, +1]$, there are 4 encoded states, as shown in Table 2.
The nonlinear mapping between quantized real numbers and their corresponding encoded states is given in Table 2.
From the table, we can see that there is a linear factor $\alpha$ between quantized real numbers and encoded states (i.e., $\alpha=3$ for Table 2). When we use Eq. (9) to compute the multiplication of two encoded vectors, the value will be expanded $\alpha ^2$ times. Therefore, the result can multiply its scaling factor to get the final result, e.g., $1/9$ as in Fig.\ 1.

\begin{table}[htbp]
\caption{Quantized real numbers and their encoded states.}
\centering
\renewcommand\arraystretch{1.5}
\begin{tabular}{p{2.5cm}<{\centering} p{1.1cm}<{\centering} p{1.1cm}<{\centering} p{1.1cm}<{\centering} p{1.1cm}<{\centering}}
\hline
\hline
Number Space & $[-1, -2/3]$ & $(-2/3, 0]$ & $(0, 2/3)$ & $[2/3, 1]$ \\
\hline
Quantized Number & $-1$ & $-1/3$ & $1/3$ & $1$ \\
\hline
Encoded States & $\{-1,-1\}$ & $\{-1,+1\}$ & $\{+1,-1\}$ & $\{+1,+1\}$ \\
\hline
\hline
\end{tabular}
\end{table}

\begin{figure*}
\setlength{\abovecaptionskip}{0.1cm}
\setlength{\belowcaptionskip}{-0.2cm}
\centering
\includegraphics[width=5.0in]{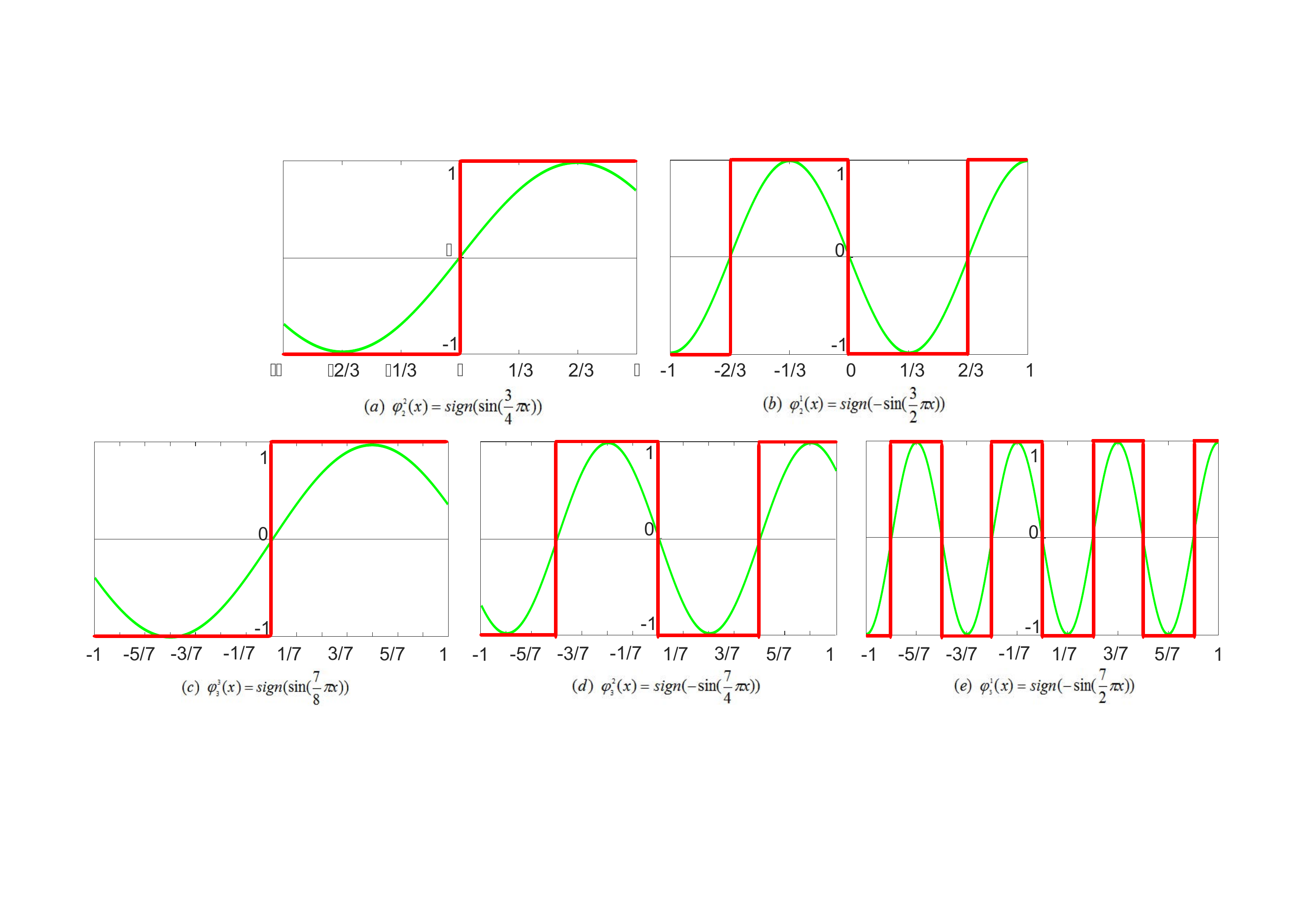}
\vspace{-1mm}

\caption{Encoding functions. (a) and (b) denote the encoding functions of the second bit and the first bit of 2BitEncoder.
(c), (d) and (e) denote the encoding functions of the third bit, the second bit and the first bit of 3BitEncoder, respectively.}
\end{figure*}

From the above results, we obviously need an activation function to achieve the transformation from the quantized number to its corresponding encoded state.
The most straightforward transformation method is to determine the \emph{encoded states} according to the number value through the look-up table, which is suitable for many hardware platforms (e.g., FPGA and ASIC).
However, its derivative is hardly be defined, and thus this method can not be applied in training process. Therefore, we need a kind of derivable encoding function as activation functions to train the network.

This is a typical single-input multiple-output problem.
The Fourier neural network (FNN) \cite{silvescu1999fourier,minami1999real} can be used to transform a nonlinear optimization problem into a combination of multiple linear ones \cite{mingo2004fourier}. It has been successfully used for stock prediction \cite{minami1999real}, aircraft engine fault diagnostics \cite{tan2006fourier}, harmonic analysis \cite{germec2009fourier},  lane departure prediction \cite{tan2017use}, and so on.
In\cite{parascandolo2016taming}, it has verified that the trigonometric activation function can achieve faster and better results than those using established monotonic functions for some certain tasks.
Fig.\ 2 shows the illustration of the $2$-bit and $3$-bit encoding functions. We can see that those encoding functions are required to be periodic, and they should have different periods. Inspired by FNN, we apply trigonometric functions as the basic encoding functions, whose lines are highlighted in red. Finally, we use the sign function to determine whether the value is $-1$ or $+1$.
Hence, we use several trigonometric functions to approximate the nonlinear mapping between quantized numbers and their encoded states, which is called as the $MBitEncoder$ function, and is formulated as follows:
\begin{eqnarray}
MBitEncoder(x)=\left\{\begin{matrix}
\!\!\! \varphi _{M}^{m}(x):sign(sin(\frac{2^M-1}{2^m}\pi \cdot x)),\; \\
 m=M\\
 \varphi _{M}^{m}(x):sign(-sin(\frac{2^M-1}{2^m}\pi \cdot x)), \\
  \quad \qquad m \in \{1,2,...,M\!-\!1\}\\
\end{matrix}\right.
\end{eqnarray}
when $\varphi _{M}^{M}(x)$ (i.e., $m\!=\!M$) is the encoding function of the highest bit of the $MBitEncoder$ function.
For example, for the 2-bit case, the $2BitEncoder$ function is formulated as follows:
\begin{eqnarray}
2BitEncoder(x)=\left\{\begin{matrix}
\varphi _{2}^{2}(x):sign(sin(\frac{3}{4}\pi \cdot x)) \quad \\
\varphi _{2}^{1}(x):sign(-sin(\frac{3}{2}\pi \cdot  x)) \quad\!\!\!\!\!
\end{matrix}\right.
\end{eqnarray}
where $\varphi _{2}^{1}(x)$ is the encoding function of the first bit ($x^i_1$) of the $2BitEncoder$ function, and $\varphi _{2}^{2}(x)$
denotes the encoder function of the second bit ($x^i_2$) of the $2BitEncoder$ function. The periodicity is clearly different from the others because it needs to denote more states.

\section{Training Networks}
To train QNNs, we usually face the problem that the derivative is not defined, and thus traditional gradient optimization methods are not applicable. \cite{Courbariaux2016Binarized} presented the \emph{HTanh} function to binarily quantize both weights and activations, and also defined the derivative to support the back-propagation (BP)  training process. They used the loss computed by binarized parameters to update full precision parameters. Similarly, \cite{Rastegari2016XNOR} also proposed to update the weights with the help of the second parameters. \cite{Bengio2013Estimating} discussed that using STE to train network models containing discrete variables can obtain faster training speed and better performance.
By utilizing equivalent transformation of our method, QNNs can be decomposed to MBBNs. The goal of training networks is to get the optimum parameters, which are quantized to $-1$ or $1$. Obviously, we have two ways to train our model: The first way is to train the decomposed MBBNs directly, and the second one is that we first to train the quantized networks to get the quantized weights, and then we use our equivalent transformation method to transform the quantized weights to the bit encoded binary weights. Next, we will describe the two ways in detail.

\subsection{Training Multi-Branch Binary Networks}
Generated by the decomposition of QNNs, MBBNs need to use $M$-bit encoding functions as activation functions to get the elements of each bit, which can be used by more efficient bitwise operations to replace arithmetic operations.
As described in Section 3.3, $M$-bit encoding functions are the combination of a series of trigonometric and sign functions.
The sign function of the encoder makes it difficult to implement the BP process. Thus, we approximate the derivative of the encoding function with respect to $x$ as follows:
\begin{eqnarray}
\left\{\begin{matrix}
\frac{\partial \varphi _{M}^{m}(x)}{\partial x}=\left\{\begin{matrix}
\frac{2^M-1}{2^m}\pi cos(\frac{2^M-1}{2^m}\pi x), & -1\leqslant x\leqslant 1, \\
0 & \textup{otherwise},
\end{matrix}\right. , \\ m=M \\
\frac{\partial \varphi _{M}^{m}(x)}{\partial x}=\left\{\begin{matrix}
-\frac{2^M-1}{2^m}\pi cos(\frac{2^M-1}{2^m}\pi x), & -1\leqslant x\leqslant 1, \\
0 & \textup{otherwise},
\end{matrix}\right. , \\ m \in \{1,2,...,M\!-\!1\} \\
\end{matrix}\right.
\end{eqnarray}

We take the 2-bit encoding as an example to describe the optimization method of MBBNs.
\begin{eqnarray}
\left\{\begin{matrix}
\frac{\partial \varphi _{2}^{2}(x)}{\partial x}=\left\{\begin{matrix}
\frac{3}{4}\pi cos(\frac{3}{4}\pi x), & -1\leqslant x\leqslant 1, \\
0, & \textup{otherwise},
\end{matrix}\right. \\

\frac{\partial \varphi _{2}^{1}(x)}{\partial x}=\left\{\begin{matrix}
-\frac{3}{2}\pi cos(\frac{3}{2}\pi x), & -1\leqslant x\leqslant 1, \\
0, & \textup{otherwise}.
\end{matrix}\right.
\end{matrix}\right.
\end{eqnarray}


Besides activations, all the weights of networks also need to be quantized to binary values.
Therefore, MBBN can be viewed as a special binarized neural network \cite{Courbariaux2016Binarized}.
In this training method, we use the same mechanisms to constrain and update parameters, and use $w\! \in\! [-1, 1]$ and $w_b \!\in\! \{-1, 1\}$ to represent the real-valued weights and its binarized weights, respectively.
$w$ is constrained between $-1$ and $1$ by the \emph{HTanh()} function to avoid excessive growth.
Instead of $M$-bit encoding functions that are complex, we use the sign function directly to binarize parameters.
Therefore, the binarization function for transforming $w$ to $w_b$ can be directly formulated as follows:
\begin{eqnarray}
w_b=Binarize(w)=sign(HTanh(w)).
\end{eqnarray}
For this function, we define the gradient function of each component to constrain the search space.
In the training process, we apply $w_b$ to compute the loss and gradients, which are used to update $w$.
That is, the input of the sign function can be constrained to $[-1, +1]$ by $HTanh(x)$, and it can also speed up convergence. The parameters of the whole network are updated in the condition of differentiability.

We give the detailed training algorithm of MBBN in \emph{Algorithm 1}, where we describe the main computing process of the network, and some details about batch normalization, pooling and loss layers are omitted. Note that the \emph{forward}() operation is the basic operation in DNNs, such as convolutional and full-connected operations. Those operations can be accelerated by binary bitwise operations.
We use \emph{backward}() and \emph{BackMBitEncoder}() to specify how to backpropagate through \emph{forward}() and \emph{MBitEncoder}() operations. \emph{Update}() specifies how to update the parameters with their known gradients. For example, we usually use \emph{Adam} or \emph{SGD} as our optimization algorithm.

\begin{algorithm}[!h]
\caption{Training an $L$-layer MBBN with $K$-bit weights and $M$-bit activations. Weights and activations are quantized according to Eqs.\ (14) and (10), respectively.}
    \hspace*{0.02in} {\bf Require:}
    A mini-batch of inputs and targets $(a_0, a^*)$, previous weights $\{W^1, W^2, ..., W^K\}$, and learning rate $\eta$.\\
    \hspace*{0.02in} {\bf Ensure:}
    Updated weights $\{W^{1*}, W^{2*}, ..., W^{K*}\}$.\\
    \hspace*{0.2in}$\{$1. Computing the parameter gradients:$\}$ \\
    \hspace*{0.2in}$\{$1.1 Forward propagation:$\}$
	\begin{algorithmic}[1]
		\FOR{$l=1$ to $L$}
        \FOR{$k=1$ to $K$}
        \STATE $W^{kb}_l \longleftarrow Bianrize(W^{k}_l)$;
        \ENDFOR
        \STATE $\hat{a}_l \longleftarrow \sum_{m=1}^{M}\sum_{k=1}^{K}2^{m+k-2}forward(a^{mb}_{l-1}, W^{kb}_l)$;
        \STATE $a_l \longleftarrow \frac{r\hat{a}_l}{(2^M-1)(2^K-1)}$;
		\IF{$l<L$}
		\STATE $a^{1b}_l, a^{2b}_l, ..., a^{Mb}_l \longleftarrow MBitEncoder(a_l)$;
		\ENDIF
        \ENDFOR

        \hspace*{0.2in}$\{$1.2 Backward propagation:$\}$  \\
        \hspace*{0.2in}Compute $g_{a_L}=\frac{\partial C}{\partial a_L}$ with the given $a_L$ and $a^*$;
        \FOR{$l=L$ to $1$}
        \STATE $g_{a_l} \longleftarrow  BackMBitEncoder(g_{a^{1b}_l}, g_{a^{2b}_l}, ..., g_{a^{Mb}_l})$;
        \STATE $g_{\hat{a}_l} \longleftarrow  \frac{r g_{a_l}}{(2^M-1)(2^K-1)}$;
        \FOR{$m=1$ to $M$}
        \STATE $g_{a^{mb}_{l-1}} \longleftarrow  \sum_{k=1}^{K}2^{m+k-2}backward(a^{mb}_{l-1}, W^{kb}_l,g_{\hat{a}_l})$;
        \ENDFOR
        \FOR{$k=1$ to $K$}
        \STATE $g_{W^{kb}_l} \longleftarrow \sum_{m=1}^{M}2^{m+k-2}backward(a^{mb}_{l-1}, W^{kb}_l, g_{a^{mb}_{l-1}})$;
        \ENDFOR
        \ENDFOR

        \hspace*{0.2in}$\{$2. Accumulating the parameter gradients:$\}$ \\
        \FOR{$l=1$ to $L$}
        \FOR{$k=1$ to $K$}
        \STATE $g_{W^{k}_l} = g_{W^{kb}_l} \frac{\partial W^{kb}_l}{\partial W^{k}_l}$;
        \STATE $W^{k*}_l \longleftarrow Update(W^{k}_l, g_{W^{k}_l}, \eta)$.
        \ENDFOR
        \ENDFOR
	\end{algorithmic}
\end{algorithm}

\subsection{Training Quantized Networks}
The above training scheme is proposed to optimize binary networks, which are transformed by our model decomposition method.
However, this decomposition method can produce multiple times as many parameters as  the original network.
If we optimize the binary network, it may easily fall into local optimal solutions and has slow convergence speed.
Based on the nonlinear mapping between quantized numbers and their encoded states, we can directly optimize the quantized network and then utilize our equivalent transformation method to decompose quantized network. The decomposed parameters can be used by MBBN in the inference process to achieve computational acceleration.

Two quantization schemes are usually applied in QNNs \cite{Courbariaux2016Binarized,Zhou2016DoReFa,miyashita2016convolutional}, and include linear quantization and logarithmic quantization.
Due to the requirement of our encoding mechanism, linear quantization is used to quantize our networks.
We quantize a tensor $x$ linearly into $K$-bit, which can be formulated as follows:
\begin{eqnarray}
Quantize_K(x) = \textrm{round}(\textrm{clamp}(x/t, 1)*(2^{K-1}-1))*d
\end{eqnarray}
where the clamp function is used to truncate all values into the range of $[-1, 1]$, $t$ is a learned parameter, and the scaling factor $d$ is defined as: $d = t/(2^{K-1}-1)$.
The rounding operation can be used to quantize a real number to a certain state, we call it a hard ladder function, which can segment input space to multi-states.
Table 2 lists the four states quantized by Eq.\ (11). However, the derivative of this function is almost zero everywhere, and thus it cannot be applied in the training process. Inspired by STE \cite{Bengio2013Estimating}, we use the same technique to speed up computing process and yield better performance. We use the loss computed by quantized parameters to update full precision parameters. Note that for our encoding scheme with low-precision quantization (e.g., binary), we use \emph{Adam} to train our model, otherwise \emph{SGD} is used.
\emph{Algorithm 2} shows the main training process of our quantized network, where
$BackQuantize_M$() specifies how to backpropagate through the $Quantize_M$() function with $M$-bit quantization.

\begin{algorithm}[!h]
	\caption{Training an $L$-layer QNN with $K$-bit weights and $M$-bit activations. Weights and activations are quantized by using Eq. (15).}
    \hspace*{0.02in} {\bf Require:}
   A mini-batch of inputs and targets $(a_0, a^*)$, previous weights $W$, and learning rate $\eta$.\\
    \hspace*{0.02in} {\bf Ensure:}
    Trained weights $W^*$. \\
    \hspace*{0.2in}$\{$1. Computing the parameter gradients:$\}$ \\
    \hspace*{0.2in}$\{$1.1 Forward propagation:$\}$
	\begin{algorithmic}[1]
		\FOR{$l=1$ to $L$}
        \STATE $W^{K}_l \longleftarrow Quantize_K(W_l)$;
        \STATE $a_l \longleftarrow forward(a^{M}_{l-1}, W^{K}_l)$;
		\IF{$l<L$}
		\STATE $a^{M}_l \longleftarrow Quantize_M(a_l)$;
		\ENDIF
        \ENDFOR

        \hspace*{0.2in}$\{$1.2 Backward propagation:$\}$  \\
        \hspace*{0.2in}Compute $g_{a_L}=\frac{\partial C}{\partial a_L}$ with the given $a_L$ and $a^*$;
        \FOR{$l=L$ to $1$}
        \STATE $g_{a_l} \longleftarrow  BackQuantize_M(g_{a^M_l})$;
        \STATE $g_{a^M_{l-1}} \longleftarrow  Backward(g_{a_l}, W^{K}_l)$;
        \STATE $g_{W^{K}_l} \longleftarrow  Backward(g_{a_l}, a^M_{l-1})$;
        \ENDFOR

        \hspace*{0.2in}$\{$2. Accumulating the parameter gradients:$\}$ \\
        \FOR{$l=1$ to $L$}
        \STATE $g_{W_l} = g_{W^{K}_l} \frac{\partial W^{K}_l}{\partial W_l}$;
        \STATE $W^{*}_l \longleftarrow Update(W_l, g_{W_l}, \eta)$;
        \ENDFOR
	\end{algorithmic}
\end{algorithm}

\section{Experiments}

In this section, we compare the performance of our method with those of some typical methods for image classification tasks on CIFAR-10 and ImageNet, object detection tasks on PASCAL VOC2007/2012, and semantic segmentation tasks on PASCAL VOC2012. In recent years, many scholars are devoted to improving the performance (e.g., accuracy and compression ratio) of QNNs, while very few researchers have studied their engineering acceleration, which is an important reason for hindering industrial promotion. Therefore, we mainly focus on an acceleration method, which is especially suitable for engineering applications.

\subsection{Image Classification}

\begin{table*}[htbp]
\centering
\caption{Classification accuracies of Lenet on CIFAR-10, and ResNet-18 and ResNet-50 on ImageNet.}
\vspace{-2mm}
\renewcommand\arraystretch{1.3}
\centering
\begin{tabular}{c |c |c |c c |c c}
\hline
\hline
 Models & & Lenet & \multicolumn{2}{c|}{ResNet-18} & \multicolumn{2}{c}{ResNet-50}\\
\hline
Methods & Compression ratio& CIFAR-10 & \!ImageNet (Top-1)\! & \!ImageNet (Top-5)\! & \!ImageNet (Top-1)\! & \!ImageNet (Top-5)\!\\
\hline
BWN \cite{Courbariaux2015BinaryConnect}& 32$\times$& 90.10\% & 60.80\% & 83.00\%& - & - \\
\hline
BNN \cite{Courbariaux2016Binarized} & 32$\times$& 88.60\% & 42.20\% & 67.10\% & - & - \\
\hline
XNOR-Net \cite{Rastegari2016XNOR} & 32$\times$& - & 51.20\% & 73.20\% & - & - \\
\hline
TWN \cite{Li2016Ternary} & 16$\times$& 92.56\% & 61.80\% & 84.20\% & - & - \\
\hline
\!\!DoReFa-Net[1/4-bit] \cite{Zhou2016DoReFa}& 32$\times$& - & 59.20\% & 81.50\% & - & - \\
\hline
ABC-Net[5-bit] \cite{lin2017towards}& 6.4$\times$& - & 65.00\% & 85.90\% & 70.10\% & 89.70\%\\
\hline
INQ[5-bit] \cite{Zhou2016Incremental}& 6.4$\times$& - & 68.92\% & 89.10\% & 74.81\% & 92.45\%\\
\hline
SYQ[2/8-bit] \cite{Faraone2018SYQ}& 16$\times$& - & 67.70\% & 87.80\% & 72.30\% & 90.90\%\\
\hline
Full-Precision & 1$\times$& 91.40\% & {70.10\%} & {89.81\%} & 76.20\% & 93.21\%\\
\hline
\hline
\multicolumn{6}{c}{Encoded activations and weights}\\
\hline
\hline
MBBN($M\!\!=\!\!K\!\!=\!\!1$) & 32$\times$ & 90.39\% & {52.33\%} & {79.51\%} & {56.65\%} & {80.78\%}\\
\hline
MBBN($M\!\!=\!\!K\!\!=\!\!2$)& 16$\times$& 91.06\% & {66.07\%} & {87.28\%} & {73.04\%} & {91.33\%}\\
\hline
MBBN($M\!\!=\!\!K\!\!=\!\!3$)& 10.7$\times$& {91.27\%} & {69.17\%} & {89.24\%} & {75.21\%} & {92.54\%}\\
\hline
MBBN($M\!\!=\!\!K\!\!=\!\!4$)& 8$\times$& 91.15\% & {70.31\%} & {89.89\%} & {76.22\%} & {93.09\%} \\
\hline
MBBN($M\!\!=\!\!K\!\!=\!\!5$)& 6.4$\times$& 90.92\% & {70.98\%} & {90.17\%} & {76.53\%} & {93.27\%} \\
\hline
MBBN($M\!\!=\!\!K\!\!=\!\!6$)& 5.3$\times$& 91.01\% & {71.42\%} & {90.51\%} & {76.61\%} & {93.44\%} \\
\hline
MBBN($M\!\!=\!\!K\!\!=\!\!7$)& 4.6$\times$& 90.20\% & {{71.46\%}} & {{90.66\%}} & {76.65\%} & {93.54\%} \\
\hline
MBBN($M\!\!=\!\!K\!\!=\!\!8$)& 4$\times$& 90.43\% & {71.46\%} & {90.54\%} & {{76.71\%}} & {{93.40\%}} \\
\hline
\hline
\end{tabular}
\end{table*}

\textbf{CIFAR-10} is an image classification dataset, which has $50,000$ training and $10,000$ testing images. All the images are $32\! \times \!32$ color images representing airplanes, automobiles, birds, cats, deer, dogs, frogs, horses, ships and trucks.

We validated our method with different bit encoding schemes, in which activations and weights are equally treated, that is, both of them use the same number of encoding bits. Table 3 lists the results of our method and many state-of-the-art models, including BWN \cite{Courbariaux2015BinaryConnect}, BNN \cite{Courbariaux2016Binarized}, XNOR-Net \cite{Rastegari2016XNOR}, XNOR-Net \cite{Rastegari2016XNOR}, TWN \cite{Li2016Ternary}, DoReFa-Net \cite{Zhou2016DoReFa}, ABC-Net \cite{lin2017towards}, INQ \cite{Zhou2016Incremental}, and SYQ \cite{Faraone2018SYQ}.
Here we used the same network architecture as in \cite{Courbariaux2015BinaryConnect} and \cite{Courbariaux2016Binarized}, except for the encoding functions. We used $\emph{HTanh}(\cdot)$ as the activation function and employed \emph{Adam} to optimize all the parameters of the networks. From all the results, we can see that the representation capabilities of our networks encoded by 1-bit or 2-bit are enough for small-scale datasets, e.g., CIFAR-10. Our method with low-precision encoding (e.g., $2$-bit) achieved nearly the same classification accuracy as its high precision versions (e.g., $8$-bit) and full-precision models, while we can attain $\sim\!16\times$ memory saving compared with its full-precision counterpart. When both activations and weights are constrained to $1$-bit, our network structure is similar to BNN \cite{Courbariaux2016Binarized}, and our method yielded even better accuracy mainly because of our proposed encoding functions.

\textbf{ImageNet:} We further compared the performance of our method with all the other methods mentioned above on the ImageNet (ILSVRC-2012) dataset \cite{Russakovsky2015ImageNet}. This dataset consists of 1K categories images, and has over $1.2$M images in the training set and $50$K images in the validation set.
Here, we used ResNet-18 and ResNet-50 to illustrate the effectiveness of our method.
and used Top-1 and Top-5 accuracies to report classification performance. For large-scale training sets (e.g., ImageNet), it usually costs plenty of time and requires sufficient computing resources for classical full-precision models. Moreover, it will be more difficult to train quantized networks, and thus the initialization of parameters is particularly important. Here, we used $\emph{HReLU}(\cdot)$ as the activation function to constrain activations. In particular, the full-precision model parameters activated by $\emph{ReLU}(\cdot)$ can be directly used as the initialization parameters for our $8$-bit quantized network. After a little number of fine-tuning, our $8$-bit quantized networks can be well-trained. Similarly, we used the $8$-bit model parameters as the initialization parameters to train $7$-bit quantized networks, and so on. There is a special case, when we use $\emph{HReLU}(\cdot)$ and the 1BitEncoder function to encode activations, and all the activations are constrained to $1$. Thus, we used $\emph{HTanh}(\cdot)$ as the activation function for $1$-bit encoding. Note that we used \emph{SGD} to optimize parameters when the number of encoding bits is not less than $3$, and the learning rate was set to $0.1$. When the number of the encoding bits is $1$ or $2$, the convergent speed of \emph{Adam} is faster than \emph{SGD}, as discussed in \cite{Courbariaux2016Binarized,Rastegari2016XNOR}.

Table 3 lists the performance (e.g., accuracies and compression ratios) of our method and many state-of-the-art models mentioned above.
Experimental results show that our method performs much better than its counterparts.
Similarly, our method with $5$-bit encoding significantly outperforms ABC-Net[$5$-bit] \cite{lin2017towards}.
Moreover, our networks can be trained in a very short time to achieve the accuracies in our experiments, and only dozens of fine-tuning are needed.
Different from BWN and TWN, whose weights are only quantized rather than activation values, our method quantifies both weights and activation values, simultaneously.
Although BWN and TWN can obtain little higher accuracies than our $1$-bit quantization model, we obtain significantly better speedups, and the speedup ratios of existing methods such as BWN and TWN are limited to $2\times$. Due to limited and fixed expression ability, existing methods (such as BWN, TWN, BNN and XNOR-Net) cannot satisfy various quantization precision requirements. In particular, our method can provide $64$ available encoding choices, and hence our encoded networks with different encoding precisions have different speedup ratios, memory requirements and experimental accuracies.

\begin{table}[htbp]
\centering
\caption{Classification accuracies of ResNet-18 with different bit precisions on ImageNet.}
\centering
\renewcommand\arraystretch{1.6}
\begin{tabular}{c |c |c |c}
\hline
\hline
Layer name & ResNet-18 & Precision\#1 & Precision\#2  \\
\hline
Conv1 & $7 \times 7$, 64, stride 2 & $M\!\!=\!\!8$, $M\!\!=\!\!8$ & $M\!\!=\!\!4$, $M\!\!=\!\!8$  \\
\hline
Conv2$\_$x & $\begin{bmatrix}
 3 \times 3, 64 \\
 3 \times 3, 64
\end{bmatrix} \times 2$ & $M\!\!=\!\!8$, $K\!\!=\!\!8$ & $M\!\!=\!\!4$, $K\!\!=\!\!5$  \\
\hline
Conv3$\_$x & $\begin{bmatrix}
 3 \times 3, 128 \\
 3 \times 3, 128
\end{bmatrix} \times 2$  & $M\!\!=\!\!8$, $K\!\!=\!\!7$ & $M\!\!=\!\!4$, $K\!\!=\!\!3$  \\
\hline
Conv4$\_$x & $\begin{bmatrix}
 3 \times 3, 256 \\
 3 \times 3, 256
\end{bmatrix} \times 2$  & $M\!\!=\!\!8$, $K\!\!=\!\!7$ & $M\!\!=\!\!4$, $K\!\!=\!\!2$  \\
\hline
Conv5$\_$x & $\begin{bmatrix}
 3 \times 3, 512 \\
 3 \times 3, 512
\end{bmatrix} \times 2$  & $M\!\!=\!\!8$, $K\!\!=\!\!6$ & $M\!\!=\!\!4$, $K\!\!=\!\!2$  \\
\hline
 FC & 1000 & $M\!\!=\!\!8$, $K\!\!=\!\!8$ & $M\!\!=\!\!4$, $K\!\!=\!\!6$  \\
\hline
\multicolumn{2}{c|}{Compression ratio} & 5.0x & 14.1x  \\
\hline
\multicolumn{2}{c|}{\multirow{2}{*}{Accuracy}} & {Top-1=71.45\%} & Top-1=69.32\%  \\
\cline{3-4}
\multicolumn{2}{c|}{} & {Top-5=90.60\%} & Top-5=89.37\% \\
\hline
\hline
\end{tabular}
\end{table}

In the above experiments, all activation values and weights are constrained to the same bit precision (e.g., $M\!\!=\!\!K\!\!=\!\!6$). However,
our method also has an advantage that it can support mixed precisions (e.g., $M\!\!=\!\!6$ and $M\!\!=\!\!5$), that is, weights and activations for each layer can have different bit precisions. This makes model quantification and acceleration be more flexible, and thus meets more requirements for model
design. We used ResNet-18 as an example to verify the effect of mixed precisions. Table 4 lists the network structures, bit precision of each layer, compression ratios, and classification accuracies on ImageNet. For the convenience, we set the same bit precision for each residual block of ResNet-18. The experimental results show that the mixed precisions can obtain higher classification accuracies.
As shown in Table 4, the classification results of \emph{Precision\#1} are higher than those of their full-precision counterparts, while obtaining a 5$\times$ compression ratio. We also set the bit precision of all the activation values to 4, and different bit precisions for each weight in Table 4, and achieve $14.1\times$ compression ratio and $69.32\%$ Top-1 accuracy.

\subsection{Object Detection}
Besides image classification, QNNs can also be used for object detection tasks \cite{wei2018quantization, li2019fully}.
Hence, we use our method to quantize some trained networks with the Single Shot MultiBox Detector (SSD) framework \cite{liu2016ssd} for object detection tasks, (i.e., a coordinate regression task and a classification task). The regression task has higher demands on the encoding precision, and thus the application of object detection presents a new challenge for QNNs.

\begin{table*}[htbp]
\large
\centering
\caption{Comparison of the object detection performance of our method with different encoding bits.}
\centering
\vspace{-2mm}
\renewcommand\arraystretch{1.2}
\resizebox{180mm}{34mm}{
\begin{tabular}{c| c| c| c c c c c c c c c c c c c c c c c c c c}
\hline
\hline
Backbone & Precision & mAP & aero & bike & bird & boat & bottle & bus & car & cat & chair & cow & table & dog & horse & mbike & person & plant & sheep & sofa & train & tv\\
\hline
\multirow{5}*{InceptionV3} & full(32-bit) & 78.8 & 80.8 & 86.8 & 81.5 & 72.1 & 53.6 & 86.9 & 87.7 & 87.4 & 62.0 & 81.4 & 75.3 & 88.0 & 87.1 & 86.0 & 79.3 & 58.7 & 81.5 & 72.9 & 87.2 & 80.2\\
\cline{2-23}
 & $M\!\!=\!\!K\!\!=\!\!8$ & 78.0 & 79.0 & 85.8 & 79.8 & 72.4 & 49.8 & 83.5 & 86.8 & 86.5 & 60.7 & 84.2 & 75.4 & 87.5 & 86.6 & 84.4 & 78.9 & 57.5 & 81.9 & 73.3 & 86.7 & 78.8\\
\cline{2-23}
 & $M\!\!=\!\!K\!\!=\!\!6$ & 77.2 & 79.1 & 85.3 & 78.9 & 67.8 & 51.4 & 85.1 & 86.4 & 87.8 & 59.3 & 82.7 & 73.3 & 87.0 & 85.5 & 83.7 & 77.7 & 56.8 & 78.8 & 70.7 & 87.9 & 78.7\\
\cline{2-23}
 & $M\!\!=\!\!K\!\!=\!\!4$ & 66.6 & 71.1 & 76.2 & 62.7 & 59.2 & 36.0 & 75.0 & 80.9 & 76.2 & 40.8 & 71.6 & 67.4 & 73.8 & 78.8 & 74.6 & 71.1 & 41.9 & 64.9 & 63.2 & 79.3 & 66.9\\
\cline{2-23}
 & BWN & 44.8 & 56.5 & 57.2 & 26.8 & 34.6 & 17.4 & 51.6 & 59.6 & 51.2 & 22.3 & 41.5 & 55.3 & 42.1 & 63.4 & 54.9 & 53.3 & 15.7 & 44.0 & 40.2 & 60.2 & 50.2\\
\hline
\multirow{5}*{ResNet-50} & full(32-bit) & 77.2 & 79.5 & 85.0 & 81.2 & 71.1 & 55.3 & 83.7 & 87.3 & 87.3 & 60.4 & 84.5 & 72.6 & 85.0 & 83.5 & 82.9 & 80.3 & 50.7 & 78.7 & 72.4 & 85.4 & 77.6\\
\cline{2-23}
 & $M\!\!=\!\!K\!\!=\!\!8$ & 78.4 & 80.7 & 86.5 & 79.8 & 70.8 & 59.4 & 85.6 & 87.5 & 88.7 & 64.1 & 84.1 & 72.1 & 85.4 & 85.4 & 86.5 & 80.6 & 53.1 & 79.1 & 73.8 & 85.4 & 78.9\\
\cline{2-23}
 & $M\!\!=\!\!K\!\!=\!\!6$ & 77.1 & 78.6 & 84.8 & 77.9 & 70.0 & 55.6 & 86.8 & 87.0 & 86.9 & 61.6 & 82.6 & 72.5 & 81.1 & 84.1 & 83.6 & 80.5 & 50.4 & 80.1 & 73.1 & 85.5 & 78.3\\
\cline{2-23}
 & $M\!\!=\!\!K\!\!=\!\!4$ & 66.4 & 70.7 & 79.2 & 58.4 & 56.9 & 41.3 & 73.5 & 81.5 & 77.0 & 47.6 & 66.6 & 67.6 & 68.4 & 77.0 & 76.6 & 73.2 & 40.0 & 64.3 & 64.1 & 77.3 & 67.4\\
\cline{2-23}
& BWN & 45.6 & 53.1 & 49.8 & 30.0 & 37.9 & 20.4 & 53.7 & 69.4 & 47.8 & 23.4 & 44.0 & 51.5 & 47.8 & 53.6 & 59.3 & 57.9 & 21.9 & 42.6 & 36.4 & 61.2 & 49.9\\
\hline
\multirow{5}*{MobileNetV2}
& full(32-bit) & 72.6 & 77.8 & 82.6 & 71.2 & 64.4 & 48.5 & 79.0 & 86.0 & 81.4 & 54.0 & 77.9 & 65.1 & 79.3 & 79.3 & 78.7 & 75.5 & 49.5 & 77.3 & 65.3 & 84.5 & 74.0\\
\cline{2-23}
 & $M\!\!=\!\!K\!\!=\!\!8$ & 71.5 & 77.3 & 81.3 & 70.8 & 64.3 & 46.0 & 77.6 & 84.5 & 79.9 & 51.9 & 75.8 & 66.5 & 78.4 & 78.4 & 78.2 & 75.6 & 47.5 & 74.0 & 68.4 & 82.2 & 70.4\\
\cline{2-23}
 & $M\!\!=\!\!K\!\!=\!\!6$ & 70.5 & 77.2 & 81.0 & 68.5 & 60.4 & 45.3 & 77.1 & 84.4 & 79.1 & 51.8 & 73.4 & 65.2 & 77.9 & 77.9 & 77.3 & 75.1 & 46.2 & 72.3 & 66.5 & 81.1 & 71.4\\
\cline{2-23}
 & $M\!\!=\!\!K\!\!=\!\!4$ & 55.1 & 68.7 & 69.6 & 42.7 & 43.9 & 27.7 & 62.9 & 76.5 & 56.9 & 29.6 & 56.2 & 56.1 & 51.8 & 51.8 & 68.1 & 64.1 & 26.5 & 55.8 & 49.6 & 72.1 & 58.4\\
\cline{2-23}
& BWN & 21.8 & 34.7 & 18.6 & 9.3 & 16.6 & 9.1 & 34.3 & 34.3 & 20.4 & 7.5 & 12.9 & 26.3 & 16.0 & 35.5 & 32.2 & 26.2 & 9.1 & 15.9 & 18.3 & 34.1 & 24.0\\
\hline
\hline
\end{tabular}
}
\end{table*}

\begin{figure*}
\centering
\setlength{\abovecaptionskip}{0.1cm}
\setlength{\belowcaptionskip}{-0.4cm}
\includegraphics[width=6in,]{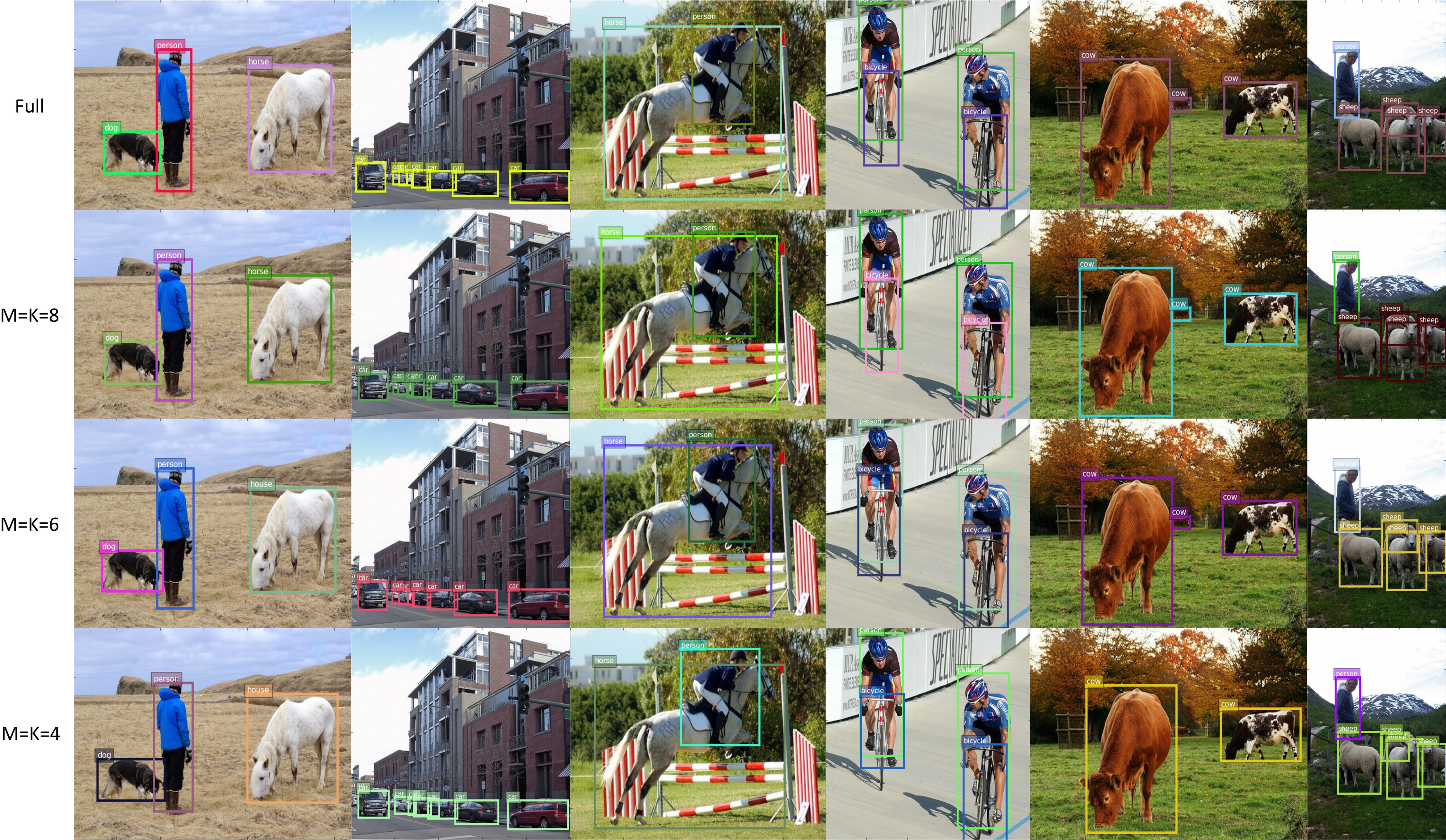}
\caption{Detection results of our method with different precisions based on the InceptionV3 network on some examples of the VOC2007 test set.}
\end{figure*}

\begin{table*}[htbp]
\large
\centering
\caption{Comparison of object detection results of our method with different encoding bits.}
\centering
\vspace{-2mm}
\renewcommand\arraystretch{1.2}
\resizebox{180mm}{11mm}{
\begin{tabular}{c| c| c| c c c c c c c c c c c c c c c c c c c c}
\hline
\hline
Backbone & Precision & mAP & aero & bike & bird & boat & bottle & bus & car & cat & chair & cow & table & dog & horse & mbike & person & plant & sheep & sofa & train & tv\\
\hline
\multirow{4}*{InceptionV3}
& $M\!\!=\!\!8$, $K\!\!=\!\!4$ & 69.9 & 70.4 & 78.0 & 68.5 & 59.8 & 40.7 & 79.2 & 82.1 & 83.3 & 48.7 & 72.8 & 74.6 & 77.6 & 80.0 & 81.1 & 72.3 & 41.4 & 67.0 & 68.2 & 80.3 & 71.3\\
\cline{2-23}
 & $M\!\!=\!\!4$, $K\!\!=\!\!8$ & 69.4 & 74.1 & 79.3 & 67.3 & 61.3 & 41.2 & 74.7 & 82.3 & 79.6 & 48.1 & 74.7 & 67.8 & 75.2 & 80.2 & 75.4 & 73.5 & 41.8 & 72.6 & 66.6 & 81.8 & 70.4\\
\cline{2-23}
 & $M\!\!=\!\!6$, $K\!\!=\!\!3$ & 67.6 & 80.8 & 86.6 & 79.7 & 73.2 & 53.8 & 84.4 & 86.9 & 87.6 & 60.4 & 83.5 & 76.3 & 87.8 & 86.2 & 85.5 & 78.1 & 56.0 & 79.9 & 70.8 & 87.0 & 77.8\\
 \cline{2-23}
 & $M\!\!=\!\!3$, $K\!\!=\!\!6$ & 64.7 & 71.1 & 74.9 & 53.8 & 55.9 & 33.7 & 72.3 & 79.4 & 77.0 & 43.2 & 66.8 & 68.8 & 69.1 & 76.1 & 74.0 & 70.3 & 37.1 & 61.7 & 67.1 & 78.4 & 64.0\\
\hline
\hline
\end{tabular}
}
\end{table*}

In this experiment, all the models were trained on the PASCAL VOC2007 and VOC2012 trainval set, and tested on the VOC2007 test set. Here, we used MobileNetV2 \cite{Sandler2018MobileNetV2}, InceptionV3 \cite{szegedy2016rethinking} and ResNet-50 \cite{he2016identity} as the basic networks to extract features in the SSD framework.
Note that we quantize the basic network layers, and keep the other layers in full-precision. All the networks are trained on a large input image of size $512\times 512$.
In order to quickly achieve convergence of the network parameters, we adopted the same training strategy in the above experiments, in which the parameters of high-bit models are used to initialize low-bit models. We used the average precision (AP) to measure the performance of the detector on each category and mean average precision (mAP) to evaluate multi-target detector performance. Table \uppercase\expandafter{\romannumeral4} shows the object detection results of our method and its counterparts, including its full-precision counterpart and BWN. It is clear that our method with $8$-bit encoding scheme can yield very similar performance as its full-precision counterparts. Especially for ResNet-50, we achieve the best results of $78.4\%$ mAP in the case of $8$-bit precision that is better than its full-precision counterpart by $1.2\%$ mAP. When we use $6$-bit to encode both activation values and weights, the evaluation index drops slightly. If the number of encode bits is constrained to $4$, the performance of this task has visibly deteriorated. Especially for MobileNetV2, when the number of the encoding bits is converted to $4$ from $6$, the mAP result drops monotonically from $70.5\%$ to $55.1\%$. For some categories (e.g., bottle, chair and plant), the AP results dramatically drop. In our experiments, we also implemented the binary weight form of different basic networks. Obviously, our method with $4$-bit encoding has a higher accuracy advantage over BWN \cite{Courbariaux2015BinaryConnect}. For object detection tasks, fixed extremely low-bit precision network seems powerless. Especially for MobileNetV2, BWN achieves the result of $21.8\%$ mAP that is not suitable for many industrial applications.

To compare the performance of different models with different precisions more conveniently, Fig. 3 shows some detection results on the examples of the VOC2007 test set. Detections are shown with scores higher than $0.5$. The detection output pictures, which are tested by different precisions, are shown in the same column. From all the results, we can see that $8$-bit and $6$-bit encoding precisions have similar performance as their full-precision counterparts. There is almost no obvious difference between those detection examples. However, when the encoding precision is constrained to $4$-bit, we can see that some offsets are produced on the bounding box (e.g., bike). Meanwhile, some small objects and obscured targets may be missed (e.g., cow and sheep).

In fact, it is not necessary to constrain both activation values and weights to the same bit precision. Table 5 shows the results of our method with four different bit precisions based on the InceptionV3 network. For the first case, we constrain activation values and weights to $8$-bit and $4$-bit, respectively. For the second case, we constrain activation values and weights to $4$-bit and $8$-bit, respectively. The two cases have the same computational complexity, however, they have different compression ratios and results.
The results of the third case, where activation values and weights are constrained to $6$-bit and $3$-bit, respectively, are obviously better than the fourth case, where activation values and weights are constrained to $3$-bit and $6$-bit, respectively. Therefore, the diversity of activation values is more important than weights. In order to mitigate the performance degradation of QNNs, we should encode activation values with a higher precision than weights.

For the object detection task, our method with a relatively low encoding precision (e.g., $6$-bit) yields good performance on the SSD framework. Similarly, it can be applied to other frameworks, e.g., R-CNN \cite{girshick2014rich}, Fast R-CNN \cite{ren2015faster}, SPP-Net \cite{he2014spatial} and YOLO \cite{redmon2016you}.

\vspace{-2mm}
\subsection{Semantic Segmentation}
As a typical task in the field of unmanned driving, image semantic segmentation has attracted many scholars and companies to invest in research.
The application scenario for this task usually relies on small devices, and therefore, the model compression and computational acceleration seems particularly important for real-world applications. We evaluate the performance of our method on the PASCAL VOC2012 semantic segmentation benchmark, which is consisted of $20$ foreground object classes and one background class. We use pixel accuracy (pixAcc) and mean IoU (mIoU) as the metrics to report the performance.

We used DeepLabV3 \cite{Chen2018DeepLab} based on ResNet-101 \cite{he2016identity} as the basic network in this experiment. Similarly, we use different encoding precisions to evaluate the semantic segmentation performance of our method. Fig.\ 5 shows the evaluating indicators of different encoding precisions, where the dotted line denotes the mIoU of the full-precision network, and the solid line denotes the pixAcc of the full-precision network. From the other curves, we can see that with the increase of encoding precisions, the semantic segmentation performance is becoming better. For instance, when the encoding precision is $7$-bit, our method achieves the best performance ($98.9\%$ mIoU and $95.2\%$ pixAcc), which is much better than its full-precision counterpart ($98.5\%$ mIoU and $92.9\%$ pixAcc). Fig.\ 4 shows some visualization results on VOC2012 with different encoding precisions. Obviously, the segmentation results can become more precise with higher encoding bits. For example, from the fourth image we can see that as the increase of encoding precisions, the detail of segmentation of horse's legs is becoming more and more clear.

In this experiment, we evaluated the performance of our method based on the DeepLabV3 framework. However, it can also be applied to FCN \cite{Long2014Fully} and PSP \cite{zhao2017pyramid}. Besides semantic segmentation tasks, our method can be applied to other real-word tasks, e.g., instance segmentation.

\begin{figure}
\setlength{\abovecaptionskip}{0.1cm}
\setlength{\belowcaptionskip}{-0.2cm}
\centering
\includegraphics[width=3.5in]{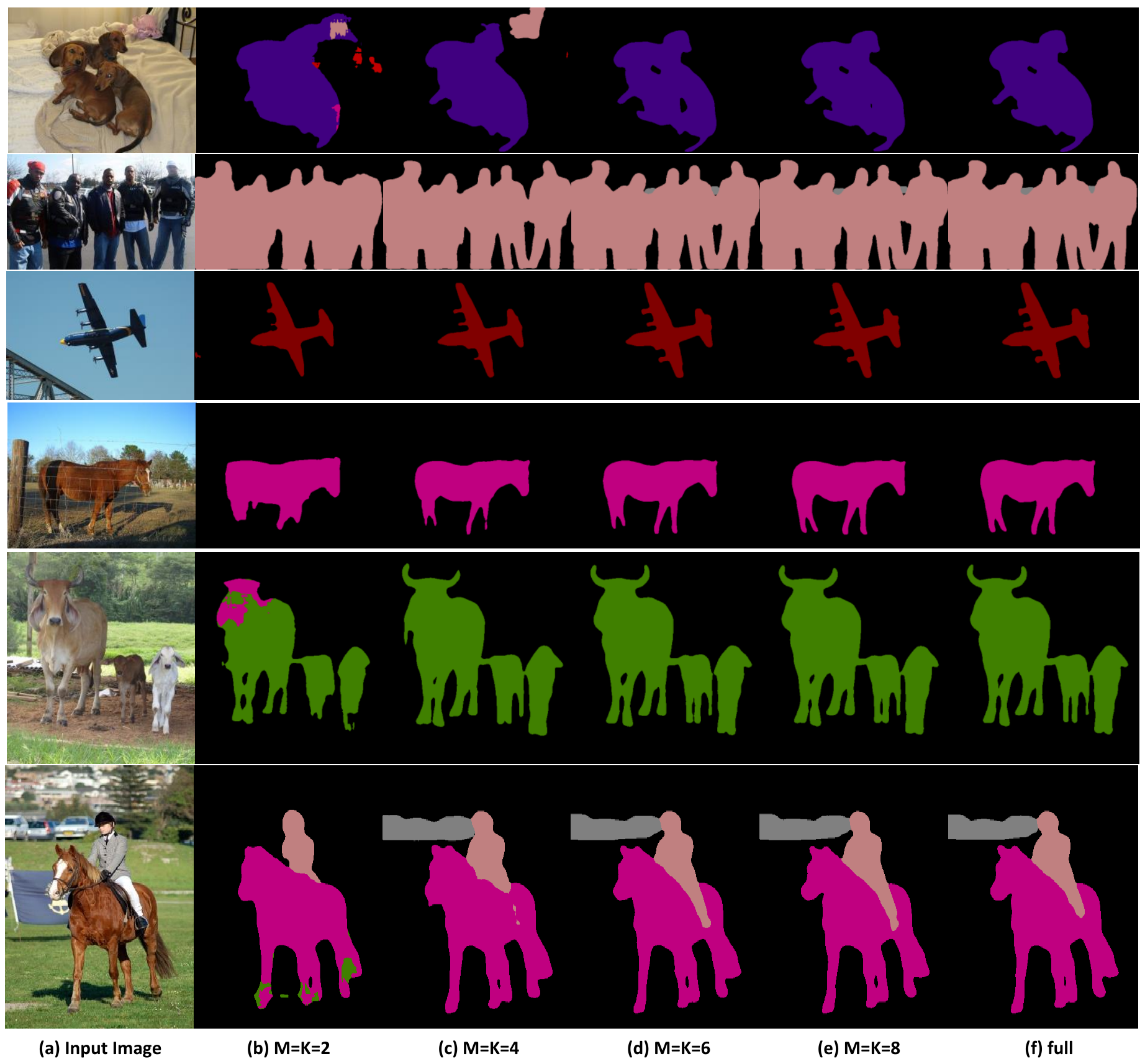}
\caption{Visualization results of our networks with different encoding precisions on the VOC2012 val set.}
\end{figure}

\begin{figure}
\setlength{\abovecaptionskip}{0.1cm}
\setlength{\belowcaptionskip}{-0.2cm}
\centering
\includegraphics[width=2.6in]{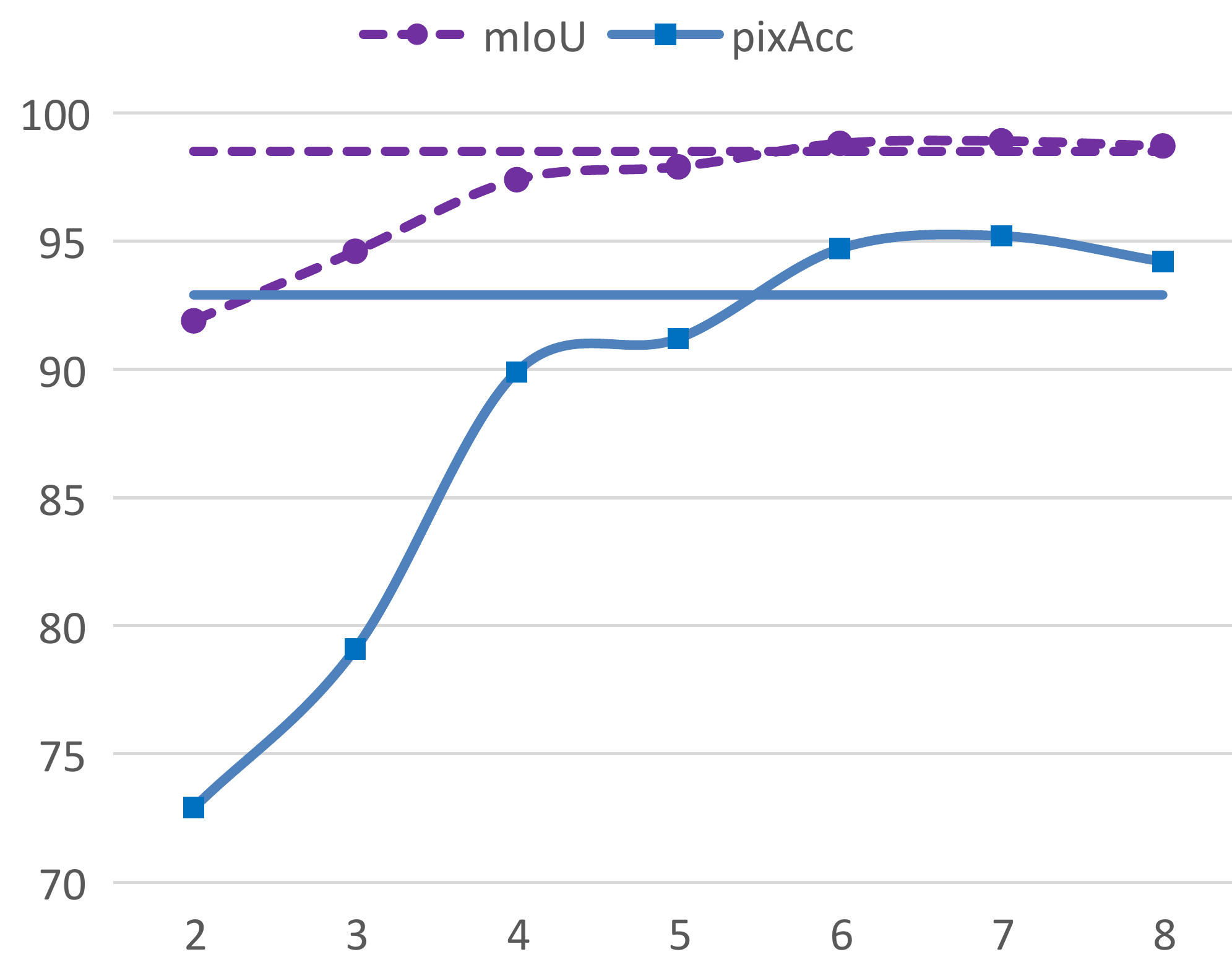}
\caption{The evaluating indicators of different encoding precisions for semantic segmentation. The dotted line and solid line denote the results of mIoU and pixAcc, respectively.}
\end{figure}

%

\vspace{-2mm}
\subsection{Efficiency Analysis}
Courbariaux \emph{et al.} \cite{Courbariaux2016Binarized} used a $8192 \times 8192 \times 8192$ matrix multiplication on a GTX750 NVIDIA GPU to illustrate the performance of XNOR kernel, and can achieve $23 \times$ speedup ratio than the baseline kernel. Therefore, we will use the same matrix multiplication to validate the performance of our method by theoretical analysis and experimental study.

In this subsection, we analyze the speed-up ratios of our method with different encoding bits. Suppose there are two vectors $x\!=\![x^1,x^2,...,x^{64}]$ and $w\!=\![w^1,w^2,...,w^{64}]$, which are composed by $64$ real numbers. If we use the $32$-bit floating-point data type to represent each element of the two vectors, the element-wise vector multiplication should require $64$ multiplications. However, after each element of the two vectors is quantized to a binary state, we use $1$-bit to represent their elements, and can use the $64$-bit data types to represent the two vectors. Then the element-wise multiplication of the two vectors can be computed at once.
This method is sometimes called SIMD (single, instuction, multiple data) within a register.
The details are shown in Fig.\ 6. The general computing platform (i.e., CPU or GPU) can perform a $64$-bit binary operation in one clock cycle. In addition, an Intel SSE (e.g., AVX or AVX-512) instruction can perform $128$ (or 256 and 512) bits binary operations. This mechanism is very suitable for the FPGA platform, which is designed for arithmetic and logical operations. Therefore, our method can achieve model compression, computational acceleration and resource saving in the inference process.

\begin{figure}
\setlength{\abovecaptionskip}{0.1cm}
\setlength{\belowcaptionskip}{-0.2cm}
\centering
\includegraphics[width=2.6in]{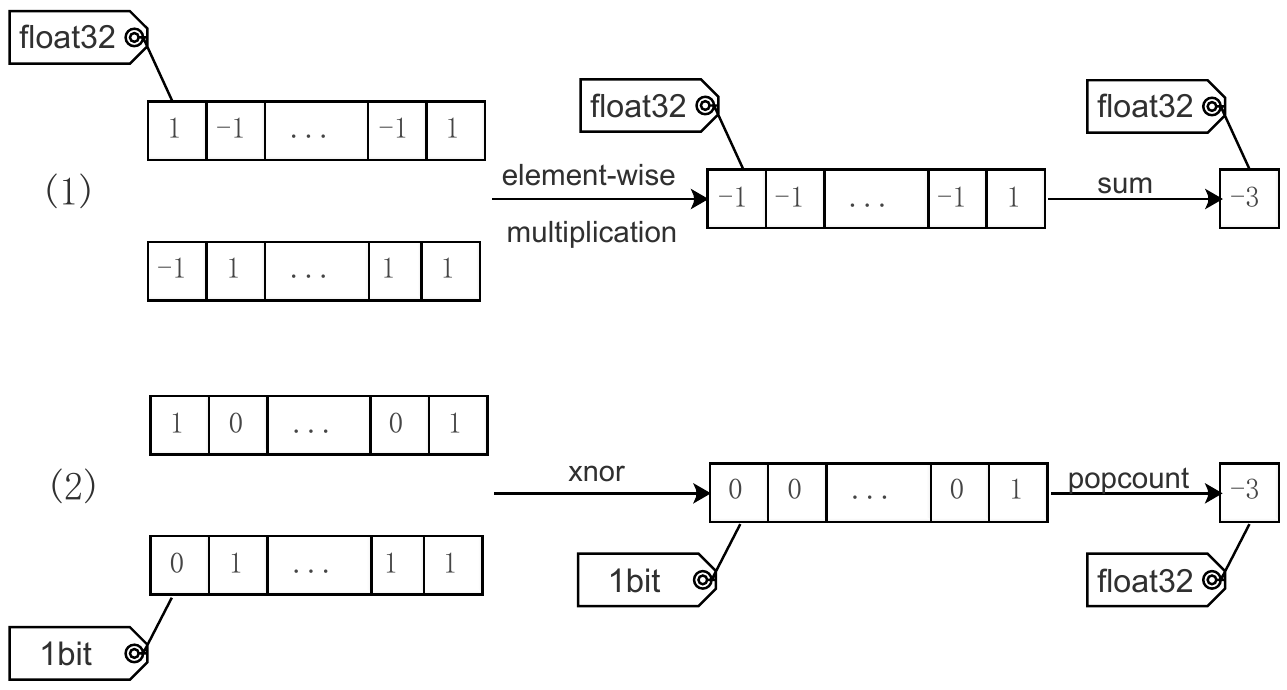}

\caption{The computing process of dot produce based on common multiplication (top) and bitwise operations (bottom).}
\end{figure}

Suppose the length of two vectors is $N$.
We use $S_{MK}$ to denote the speedup ratio of our method by constraining activation values to $M$-bit and weights to $K$-bit, where $M, K \!\in\! \{1,2,...,8\}$, and $S_{MK}$ can be formulated as follows:
\begin{eqnarray}
S_{MK} = \frac{N\gamma}{MK(\gamma+2\lceil \frac{N}{L}\rceil)+(MK-1)\beta}
\end{eqnarray}
where $\gamma$ is the speedup ratio between the multiply-accumulate operation (MAC) and bitwise operations, $\beta$ denotes the speedup ratio between addition compared and bitwise operations, and $L$ represents the length of register (e.g., 8, 16, 32, and 64). If we use $O(1)$ to represent the time complexity of $1$-bit operations, we can use $O(MK)$ to denote the time complexity of our method by constraining activation values to $M$-bit and the values of weights to $K$-bit.
Following TBN \cite{wan2018tbn}, we can know that $\gamma\!=\!1.91$ and $L\!=\!64$. According to the speedup ratio achieved by BWN \cite{Courbariaux2015BinaryConnect}, we can safely assume $\beta\!=\!\gamma/2$. Thus, the matrix multiplication after encoded by $2$-bit can obtain at most $\sim\!\!15.13\times$ speedups over its full-precision counterpart.

\begin{figure}
\setlength{\abovecaptionskip}{0.1cm}
\setlength{\belowcaptionskip}{-0.2cm}
\centering
\includegraphics[width=2.6in]{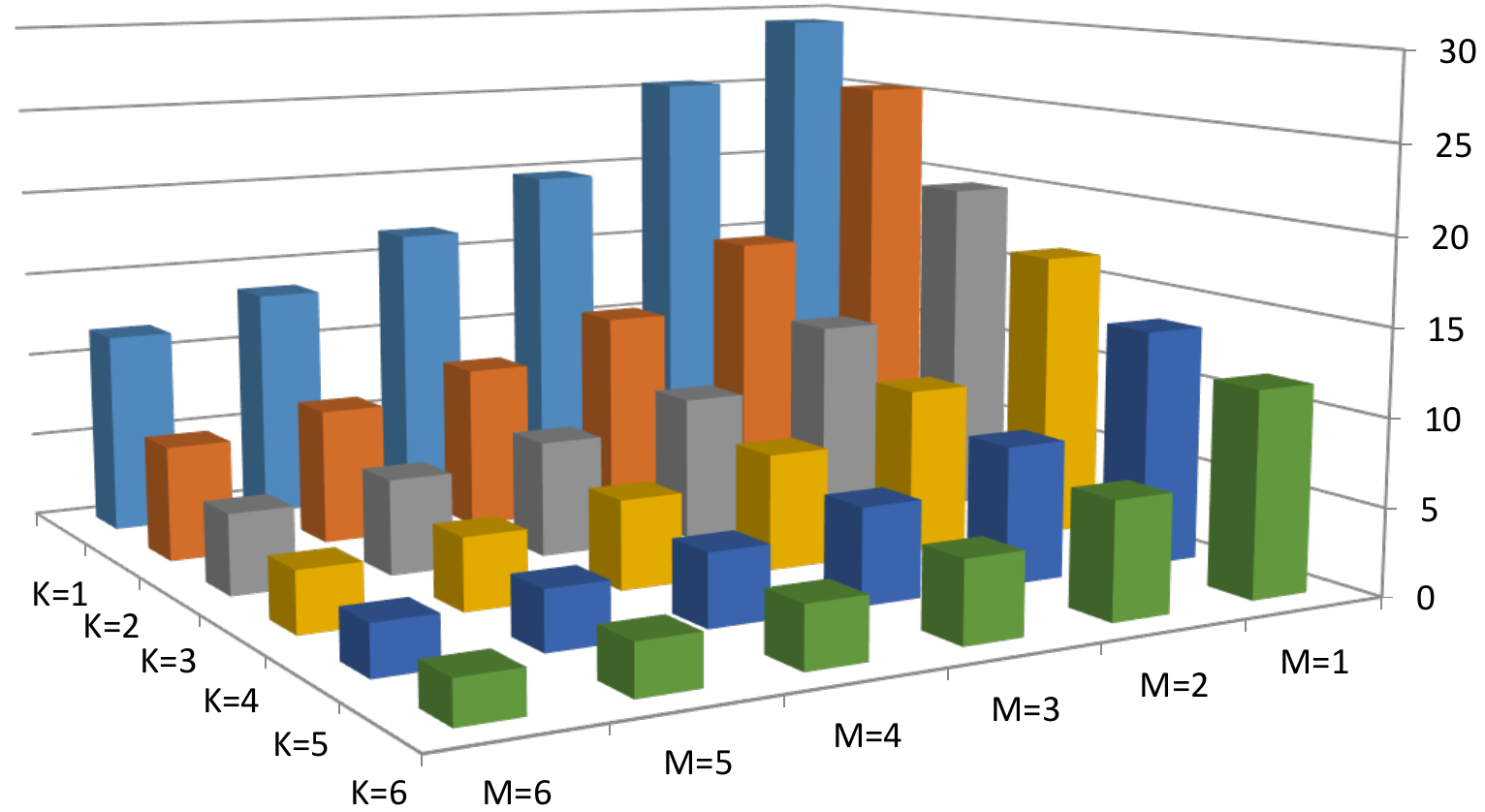}

\caption{The speedup ratios of our networks with different encoding precisions (from 1-bit to 6-bit).}
\end{figure}

Besides the theoretical analysis, we also implemented the multiplication of two matrices via our $\{-1, +1\}$ encoding scheme on GTX 1080 GPU.
Please refer to \cite{Courbariaux2016Binarized} for the experimental design and analysis.
In our experiments, we first get the encoding numbers of two matrices, and then store those numbers by bits. The matrix multiplication after encoded by $2$-bit has obtained at most $\sim\!15.89\times$ speedup ratio than the baseline kernel (a quite unoptimized matrix multiplication kernel). Because of the parallel architecture of GPU, the experimental result is slightly higher than its theoretical speedup ratio. Fig.\ 7 shows the speedup ratios of different encoding precisions (from $1$-bit to $6$-bit), where both $M$ and $K$ denote the number of encoding bits for weights and activations, respectively. From all the results, we can see that the computational complexity is gradually increasing with the increase of encoding bits. Thus, users can easily achieve different encoding precisions arbitrarily according to their requirements (e.g., accuracy and speed) and hardware resources (e.g., memory).
The code can be available at: \href{https://github.com/qigongsun/BMD}{https://github.com/qigongsun/BMD}.

\vspace{-2mm}
\section{Discussion and Conclusion}

\subsection{$\{0, 1\}$ Encoding and $\{-1, +1\}$ Encoding Schemes}
As described in \cite{Zhou2016DoReFa}, there exists a nonlinear mapping between quantized numbers and their encoded states. The quantized values are usually restricted to a closed interval $[-1, 1]$. For example, the mapping is formulated as follows:
\begin{eqnarray}
\mathrm{x}^q=\frac{2}{2^M-1}\mathrm{x}^{\{0,1\}}-1
\end{eqnarray}
where $\mathrm{x}^q$ denotes a quantized number, and $\mathrm{x}^{\{0, 1\}}$ denotes the fixed-point integer encoded by $0$ and $1$. We use a $K$-bit fixed-point integer to represent a quantized number $\mathrm{w}^q$. The product can be formulated as follows:
\begin{eqnarray}\label{equ09}
\mathrm{x}^q\cdot \mathrm{w}^q=\frac{4}{(2^M-1)(2^K-1)}\mathrm{x}^{\{0, 1\}}\cdot \mathrm{w}^{\{0, 1\}}- \qquad \nonumber \\
\qquad \frac{2}{2^M-1}\mathrm{x}^{\{0, 1\}}-\frac{2}{2^K-1}\mathrm{w}^{\{0, 1\}}+1.
\end{eqnarray}
As shown in (\ref{equ09}), the product form is a polynomial, which has four terms. Note that each term has its own scaling factor. The computation of $\mathrm{x}^{\{0,1\}}\cdot \mathrm{w}^{\{0, 1\}}$ can be accelerated by bitwise operations. However, the polynomial and scaling factors increase the computational complexity.

For our proposed quantized binary encoding scheme, the product of $\mathrm{x}^q$ and $\mathrm{w}^q$ is defined as
\begin{eqnarray}
\mathrm{x}^q\cdot \mathrm{w}^q=\frac{1}{(2^M\!-\!1)(2^K\!-\!1)}\mathrm{x}^{\{-1, +1\}}\cdot \mathrm{w}^{\{-1, +1\}}
\end{eqnarray}
where $\mathrm{x}^{\{-1, +1\}}$ and $\mathrm{w}^{\{-1, +1\}}$ denote the fixed-point integers encoded by $-1$ and $+1$. Obviously, compared with the above encoding of $\{0, 1\}$, the product can be more efficiently calculated by using our proposed encoding scheme.

\vspace{-2mm}
\subsection{Linear Approximation and Quantization}
As described in \cite{lin2017towards,Guo2017Network,Xu2018Alternating}, the weight $\mathrm{w}$ can be approximated by the linear combination of $K$ binary subitems \{$\mathrm{w}_1, \mathrm{w}_2,..., \mathrm{w}_K$\} and $\mathrm{w}_i\in \{-1, +1\}^N$, which can be computed by more efficient bitwise operations. In order to obtain the combination, we need to solve the following problem
\begin{eqnarray}
\min_{\{\alpha_i,\mathrm{w}_i \}_{i=1}^K}\left \| \mathrm{w}-\sum_{i=1}^{K}\alpha_i \mathrm{w}_i\right \|^2,  \;\mathrm{w}\in \mathbb{R}^N.
\end{eqnarray}
When this approximation is used in DNNs, $\mathrm{w}_i$ can be considered as model weights. However, the scaling factors $\alpha_i$ are introduced into this approximation, and such a scheme expands $K$ times the number of parameters. Therefore, this approximation can convert the original model into a complicated binary network, which leads to be hard to train \cite{li2017training} and easily falls into local optimal solutions.

For our method, we use the quantized parameters $\mathrm{w}^q$ to approximate $\mathrm{w}$ as follows:
\begin{eqnarray}
\mathrm{w}\approx \frac{1}{2^K-1}\mathrm{w}^q,\;\mathrm{w}\in [-1, +1]^N
\end{eqnarray}
where $\mathrm{w}^q$ is a positive or negative odd number, and its absolute value is not larger than $2^K\!-\!1$. Unlike the above linear approximation, our method can achieve the quantized weights, and directly get the corresponding encoding elements. Thus, our networks can be more efficiently trained via our quantization scheme than the linear approximation.

\subsection{Conclusions}
In this paper, we proposed a novel encoding scheme using $\{-1, +1\}$ to decompose QNNs into multi-branch binary networks, in which we used bitwise operations (\emph{xnor} and \emph{bitcount}) to achieve model compression, acceleration and resource saving.
In particular, users can easily achieve different encoding precisions arbitrarily according to their requirements (e.g., accuracy and speed) and hardware resources (e.g., memory). This special data storage and calculation mechanism can yield great performance in FPGA and ASIC, and thus our mechanism is feasible for smart chips. Our future work will focus on improving the hardware implementation, and exploring some ways (e.g., neural architecture search \cite{zoph2016neural, elsken2018neural, liu2018darts}) to automatically select proper bits for various network architectures, e.g., VGG \cite{Karen2014} and ResNet \cite{Kaiming2016}.

\vspace{-1mm}
%

\ifCLASSOPTIONcaptionsoff
  \newpage
\fi
\bibliographystyle{IEEEtran}
\bibliography{egbib}

\end{document}